\documentclass[utf8]{frontiersSCNS} 

\usepackage{url,hyperref,lineno,microtype}
\usepackage[onehalfspacing]{setspace}

\usepackage{algorithm}
\usepackage[noend]{algpseudocode}

\algnewcommand\algorithmicforeach{\textbf{for each}}
\algdef{S}[FOR]{ForEach}[1]{\algorithmicforeach\ #1\ \algorithmicdo}

\algnewcommand\algorithmicinput{\textbf{Input:}}
\algnewcommand\algorithmicoutput{\textbf{Output:}}
\algnewcommand\Input{\item[\algorithmicinput]}%
\algnewcommand\Output{\item[\algorithmicoutput]}%

\usepackage{amssymb}
\usepackage{amsmath}
\usepackage{array}
\usepackage{multirow}

\usepackage[binary-units=true]{siunitx}

\usepackage{xcolor}

\DeclareGraphicsExtensions{.pdf,.jpeg,.png}

\usepackage{textcomp}

\newcommand{\predicate}[1]{{\fontfamily{cmtt}\selectfont \small{#1}}}

\usepackage{amsthm}
\usepackage{scrextend}
\newtheoremstyle{style}
  {\topsep} 
  {\topsep} 
  {\addtolength{\leftskip}{1em}\addtolength{\rightskip}{1em}} 
  {.0em} 
  {\bfseries} 
  {:} 
  {.3em} 
  {} 
\theoremstyle{style} \newtheorem{example}{Example}

\usepackage{listings}
\usepackage[T1]{fontenc}
\usepackage{xspace}

\usepackage{subcaption,graphicx}
\usepackage{caption}

\newcommand{\circled}[1]{{\Large \textcircled{\small #1.}}}

\newcommand\compactdots{\hbox to 0.75em{.\hss.\hss.}}

\def\keyFont{\fontsize{8}{11}\helveticabold }
\def\firstAuthorLast{Zuidberg Dos Martires {et~al.}} 
\def\Authors{Pedro Zuidberg Dos Martires\,$^{1,*}$, Nitesh Kumar\,$^{1}$, Andreas Persson\,$^{2,*}$, \\ Amy Loutfi\,$^{2}$, Luc De Raedt\,$^{1}$}

\def\Address{ $^{1}$Declaratieve Talen en Artificiele Intelligentie (DTAI), Department of Computer Science, KU Leuven, 3001 Heverlee, Belgium \\
$^{2}$Center for Applied Autonomous Sensor Systems (AASS), Dept. of Science and Technology, {\"O}rebro University, 701 82 {\"O}rebro, Sweden
}

\def\corrAuthor{Pedro Zuidberg Dos Martires}
\def\corrEmail{pedro.zudo@kuleuven.be}

\begin{document}
\onecolumn
\firstpage{1}

\title[Symbolic Learning and Reasoning]{Symbolic Learning and Reasoning\\with Noisy Data for Probabilistic Anchoring}

%
\author[\firstAuthorLast ]{\Authors} 
\address{} 
\correspondance{} 
\extraAuth{Andreas Persson \\  andreas.persson@oru.se}

\maketitle

\begin{abstract} 

Robotic agents should be able to learn from sub-symbolic sensor data, and at the same time, be able to reason about objects and communicate with humans on a symbolic level. This raises the question of how to overcome the gap between symbolic and sub-symbolic artificial intelligence.
We propose a  semantic world modeling approach based on bottom-up object anchoring using an object-centered representation of the world. Perceptual anchoring processes continuous perceptual sensor data and maintains a correspondence to a symbolic representation.
We extend the definitions of anchoring  to handle multi-modal probability distributions and  we couple the resulting symbol anchoring system to a probabilistic logic reasoner for performing inference.
Furthermore, we use statistical relational learning to enable the anchoring framework to learn  symbolic knowledge in the form of a set of probabilistic logic rules of the world from noisy and sub-symbolic sensor input.
The resulting framework, which combines perceptual anchoring and statistical relational learning, is able to maintain a semantic world model of all the objects that have been perceived over time, while still exploiting the expressiveness of logical rules to reason about the state of objects which are not directly observed through sensory input data.
To validate our approach we demonstrate, on the one hand, the ability of our system to perform probabilistic reasoning over multi-modal probability distributions, and on the other hand, the learning of probabilistic logical rules from anchored objects produced by perceptual observations. The learned logical rules are, subsequently, used to assess our proposed probabilistic anchoring procedure. We demonstrate our system in a setting involving object interactions where object occlusions arise and where probabilistic inference is needed to correctly anchor objects.

\tiny
 \keyFont{ \section{Keywords:} Semantic World Modeling, Perceptual Anchoring, Probabilistic Anchoring, Statistical Relational Learning, Probabilistic Logic Programming, Object Tracking, Relational Particle Filtering, Probabilistic Rule Learning.}
\end{abstract}

\section{Introduction} 
\label{section:introduction}
\textit{Statistical Relational Learning} (SRL)~\citep{Getoor:2007:ISR:1296231,raedt2016statistical} 
tightly integrates predicate logic with graphical models in order to
extend the expressive power of graphical models towards relational logic and to
obtain probabilistic logics than can deal with uncertainty. 
After two decades of research, a plethora of expressive probabilistic logic reasoning languages and systems exists, (e.g.~\cite{sato2001parameter,richardson2006markov,Getoor.2013.PSL,fierens2015inference}). 
One obstacle that still lies ahead in the field of SRL (but see \cite{gardner2014incorporating} and \cite{beltagy2016representing}), is to combine  symbolic reasoning and learning, on the one hand, with sub-symbolic data and perception, on the other hand. 
The question is how to create a symbolic representation of the world from sensor data in order to reason and ultimately plan in an environment riddled with uncertainty and noise. In this paper,
we will take a probabilistic logic approach to study this problem in the context of perceptual anchoring.

An alternative to using SRL or probabilistic logics would be to resort to deep learning. 
Deep learning 
is based on \textit{end-to-end learning} (e.g.,~\cite{silver2016mastering}). Although exhibiting impressive results,  deep neural networks do suffer from certain drawbacks. As opposed to probabilistic rules, it is, for example, not straightforward to include prior (symbolic) knowledge in a neural system. Moreover, it is also often difficult to give guarantees for the behavior of neural systems, cf. the debate around safety and explainability in AI~\citep{huang2017safety,gilpin2018explaining}. Although not free from this latter shortcoming, this is less of a concern for symbolic systems, which implies that bridging the symbolic/sub-symbolic gap is therefore paramount. A notion that aims to bridge the symbolic/sub-symbolic gap is the definition of \textit{perceptual anchoring}, as introduced by \cite{coradeschi&saffiotti-2000,coradeschi&saffiotti-2001}. Perceptual anchoring tackles the problem of creating and maintaining, in time and space, the correspondence between symbols and sensor data that refer to the same physical object in the external world (a detailed overview of perceptual anchoring is given in Section~\ref{section:background_anchoring}). In this paper, we particularly emphasize sensor-driven \textit{bottom-up anchoring}~\citep{loutfi.et.al-2005}, whereby the anchoring process is triggered by the sensory input data. 

A further complication in robotics, and perceptual anchoring more specifically, is the inherent dependency on time. This means that a probabilistic reasoning system should incorporate the concept of time natively. One such system, rooted in the SRL community, is the probabilistic logic programming language \textit{Dynamic Distributional Clauses} (DDC)~\citep{nitti2016learning}, which can perform probabilistic inference over logic symbols and over time. In our previous work, we coupled the probabilistic logic programming language DDC to a perceptual anchoring system~\citep{persson2019semantic}, which endowed the perceptual anchoring system with probabilistic reasoning capabilities.
A major challenge in combining perceptual anchoring with a high-level probabilistic reasoner, and which is still an open research question, is the administration of \textit{multi-modal} probability distributions in anchoring\footnote{A multi-modal probability distribution is a continuous probability distribution with strictly more than one local maximum. The key difference to a uni-modal probability distribution, such as a simple normal distribution, is that summary statistics do not adequately mirror the actual distribution. In perceptual anchoring these multi-modal distributions do occur, especially in the presence of object occlusions, and handling them appropriately is critical for correctly anchoring objects. This kind of phenomenon is well known when doing filtering and is the reason why particle filters can be preferred over Kalman filters.}. In this paper, we extend the anchoring notation in order to handle additionally multi-modal probability distributions.
A second point that we have not addressed in~\cite{persson2019semantic}, is the learning of probabilistic rules that are used to perform probabilistic logic reasoning. We show that, instead of hand-coding these probabilistic rules, we can adapt existing methods present in the body of literature of SRL to learn them from raw sensor data. In other words, instead of providing a model of the world to a  robotic agent, it learns this model in form of probabilistic logical rules. These rules are then used by the robotic agent to reason about the world around it, i.e. perform inference.

In~\cite{persson2019semantic}, we showed that enabling a perceptual anchoring system to reason further allows for correctly anchoring objects under object occlusions. We borrowed the idea of encoding a \textit{theory of occlusion} as a probabilistic logic theory from~\cite{nitti2014relational} (discussed in more detail in Subsection~\ref{sec:theory_occlusion}). While~\citeauthor{nitti2014relational} operated in a strongly simplified setting, by identifying objects with AR tags, we used a perceptual anchoring system instead --- identifying objects from raw RGB-D sensor data. In contrast to the approach presented here, the theory of occlusion was not learned but hand-coded in these previous works and did not take into account the possibility of multi-modal probability distributions. We evaluate the extensions of perceptual anchoring, proposed in this paper, on three showcase examples, which exhibit exactly this behavior: 1) we perform probabilistic perceptual anchoring when object occlusion induces a multi-modal probability distributions, and 2) we perform probabilistic perceptual anchoring with a learned theory of occlusion.

We structure the remainder of the paper as follows. In Section~\ref{section:preliminaries}, we introduce the preliminaries of this work by presenting the background and motivation of used techniques.
Subsequently, we discuss, in Section \ref{section:anchoring}, our first contribution by first giving a more detailed overview of our prior work~\citep{persson2019semantic}, followed by introducing a probabilistic perceptual anchoring approach in order to enable anchoring in a multi-modal probabilistic state-space. We continue, in Section~\ref{section:learning_ddc}, by explaining how probabilistic logical rules are learned. In Section \ref{section:experiments}, we evaluate both our contributions on representative scenarios before closing this paper with conclusions, presented in Section~\ref{section:conclusions}.

\section{Preliminaries}\label{section:preliminaries}

\subsection{Perceptual Anchoring}
\label{section:background_anchoring}

Perceptual anchoring, originally introduced by \cite{coradeschi&saffiotti-2000,coradeschi&saffiotti-2001},  addresses a subset of the symbol grounding problem in  robotics and intelligent systems. The notion of perceptual anchoring has been extended and refined since its first definition. Some notable refinements include the integration of \textit{conceptual spaces}~\citep{chella.et.al-2003, chella.et.al-2004}, the addition of \textit{bottom-up anchoring}~\citep{loutfi.et.al-2005}, extensions for \textit{multi-agent systems}~\citep{leblanc&saffiotti-2008}, considerations for non-traditional sensing modalities and \textit{knowledge based anchoring} given full scale knowledge representation and reasoning systems~\citep{loutfi-2006, loutfi&coradeschi-2006, loutfi.et.al-2008}, and \textit{perception and probabilistic anchoring}~\citep{blodow.et.al-2010}. All these approaches to perceptual anchoring share, however, a number of common ingredients from~\cite{coradeschi&saffiotti-2000,coradeschi&saffiotti-2001}, including: 
\begin{itemize}
    \item A \textit{symbolic system} (including: a set $\mathcal{X} = \{ x_1, x_2,\dots \}$ of \textit{individual symbols}; a set $\mathcal{P} = \{ p_1, p_2,\dots \}$ of \textit{predicate symbols}).
    \item A \textit{perceptual system} (including: a set $\Pi = \{ \pi_1, \pi_2,\dots \}$ of \textit{percepts}; a set $ \Phi = \{ \phi_1, \phi_2,\dots \}$ of \textit{attributes} with values in the domain $D(\phi_i)$).
    \item \textit{Predicate grounding relations} $g \subseteq \mathcal{P} \times \Phi \times D(\Phi)$ that encode the correspondence between unary predicates and values of measurable attributes (i.e., the relation $g$ maps a certain predicate to compatible attribute values). 
\end{itemize}

While the traditional definition of \cite{coradeschi&saffiotti-2000,coradeschi&saffiotti-2001} assumed \textit{unary} encoded perceptual-symbol correspondences,  this does not support 
the maintenance of anchors with different attribute values at different times. To address this problem, \cite{persson.et.al-2017}   distinguishes  two different types of attributes:
\begin{itemize}
    \item \emph{Static attributes} $\phi$, which are unary within the anchor according to the traditional definition.
	\item \emph{Volatile attributes} $\phi_t$, which are individually indexed by time $t$, which are maintained in a set of attribute instances $\varphi$, such that $\phi_t \in \varphi$.
\end{itemize}

Without loss of generality, we assume from here on that all \textit{attributes stored in an anchor} are volatile, i.e., that they are indexed by a time step $t$. Static attributes are trivially converted to volatile attributes by giving them the same attribute value in each time step.

\begin{figure}[ht!]
	\begin{center}
		\includegraphics[width=0.84\textwidth]{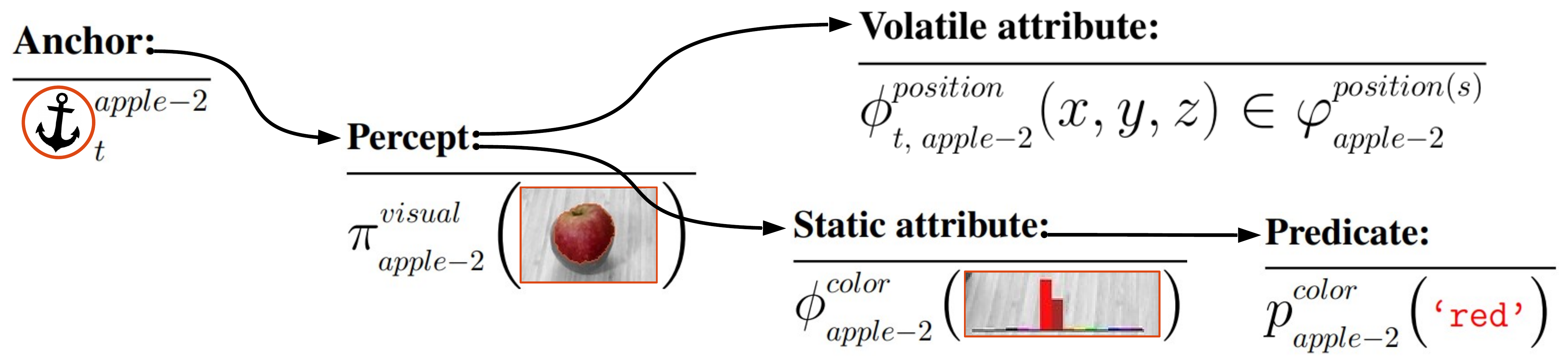}
	\end{center}
	\caption{A conceptual illustration of the internal data structure that constitutes a single anchor, and which is first initiated by a percept $\pi$ from a raw image. The volatile and static attributes are derived from this percept, while predicates such as \predicate{red}, are derived from static attributes (which are not indexed by time), e.g. the static color histogram attribute.}
	\label{fig:conceptual_anchoring} 
\end{figure}

Given the components above, an \textit{anchor} is an internal data structure $\alpha^x_t$, indexed by time $t$ and identified by a unique individual symbol $x$ (e.g., \predicate{mug-1} and \predicate{apple-4}), which encapsulates and maintains the correspondences between percepts and symbols that refer to the \textit{same physical object}, as depicted in Figure~\ref{fig:conceptual_anchoring}. Following the definition presented by \cite{loutfi.et.al-2005}, the principal functionalities to create and maintain anchors in a \textit{bottom-up fashion}, i.e., functionalities triggered by a perceptual event, are:

\begin{itemize}
 
	\item \textit{Acquire} -- initiates a new anchor whenever a candidate object is received that does not match any existing anchor $\alpha^x_t$. This functionality defines a structure $\alpha^x_t$, indexed by time $t$ and identified by a unique identifier $x$, which encapsulates and stores all perceptual and symbolic data of the candidate object. 

	\item \textit{Re-acquire} -- extends the definition of a matching anchor $\alpha^x_t$ from time $t-k$ to time $t$. This functionality ensures that the percepts pointed to by the anchor are the most recent and adequate perceptual representation of the object.
 
\end{itemize}

Based on the functionalities above, it is evident that an \textit{anchoring matching function} is essential to decide whether a candidate object is matches an existing anchor or not. Different approaches in perceptual anchoring vary, in particular in how the matching function is specified. For example, in~\cite{persson2019semantic}, we have shown that the anchoring matching function can be approximated by a learned model trained with manually labeled samples collected through an \textit{annotation interface} (through which the human user can interfere with the anchoring process and provide feedback about which objects in the scene  match previously existing anchors).

In another recently published anchoring approach, \cite{ruiz-sarmiento.et.al-2017} focus on spatial features and distinguish \textit{unary} object features, e.g., the position of an object, from \textit{pairwise} object features, e.g., the distance between two objects, in order to build a graph-based world model that can be exploited by a \textit{probabilistic graphical model}~\citep{koller-2009} in order to leverage contextual relations between objects to support $3{\text -}D$ object recognition. In parallel with our previous work on anchoring, \cite{gunther.et.al-2018} have further exploited this graph-based model on spatial features and propose, in addition, to learn the matching function through the use of a \textit{Support Vector Machine} (trained on samples of object pairs manually labeled as \textit{"same or different object"}), in order to approximate the \textit{similarity} between two objects. The assignment of candidate objects to existing anchors is, subsequently, calculated using prior similarity values and a \textit{Hungarian} method~\citep{kuhn-1955}. However, in contrast to~\cite{gunther.et.al-2018}, the matching function introduced in \cite{persson2019semantic} do not only rely upon spatial features (or attributes), but can also take into consideration visual features (such as color features), as well as semantic object categories, in order to approximate the anchoring matching problem.

\subsection{Dynamic Distributional Clauses}
\label{section:overview_ddc}
\textit{Dynamic Distributional Clauses} (DDC)~\citep{nitti2016learning} provide a framework for probabilistic programming  that extends the logic programming language \textit{Prolog}~\citep{sterling1994art} to the probabilistic domain. A comprehensive treatise on the field of probabilistic logic programming can be found in~\cite{de2015probabilistic} and \cite{riguzzi2018foundations}. DDC is capable of representing discrete and continuous random variables and to perform probabilistic inference. Moreover, DDC explicitly models time, which makes it predestined to model dynamic systems. The underpinning concepts of DDC are related to ideas presented in~\cite{milch20071} but embedded in  logic programming.
Related ideas of combining discrete time steps, Bayesian learning and logic programming are also presented in~\cite{angelopoulos2008bayesian,angelopoulos2017distributional}.

An {\em atom} $p(t_1, ..., t_n)$ consists of a predicate $p/n$ of arity $n$ and terms $t_1, ..., t_n$. A {\em term} is either a constant (written in lowercase), a variable (in uppercase), or a function symbol.  A {\em literal} is an atom or its negation. Atoms which are negated are called {\em negative atoms} and atoms which are not negated are called {\em positive atoms}.

A distributional clause is of the form $\mathtt{h}\sim \mathcal{D}\leftarrow \mathtt{b}_1,\dots,\mathtt{b}_n$,
where $\sim$ is a predicate in infix notation and $\mathtt{b}_i$'s are literals, i.e., atoms or their negation. $\mathtt{h}$ is a term representing a random variable and $\mathcal{D}$ tells us how the random variable is distributed. The meaning of such a clause is that each grounded instance of a clause $(\mathtt{h}\sim \mathcal{D}\leftarrow \mathtt{b}_1,\dots,\mathtt{b}_n)\theta$ defines a random variable $\mathtt{h}\theta$ that is distributed according to $\mathcal{D}\theta$ whenever all literals  $b_i\theta$ are true. A grounding substitution $\theta=\{ \mathtt{V}_1 / \mathtt{t}_1, \dots, \mathtt{V}_n / \mathtt{t}_n \}$ is a transformation that simultaneously substitutes all logical variables $\mathtt{V}_i$ in a distributional clause with non-variable terms $\mathtt{t}_i$.
DDC can be viewed as a language that defines conditional probabilities for discrete and continuous random variables:
$p(\mathtt{h}\theta | \mathtt{b}_1 \theta, \dots, \mathtt{b}_n \theta)=\mathcal{D}\theta$.

\begin{example}\label{example:ddc_example}
Consider the following DDC program:

\begin{addmargin}{1em}

{\small
\begin{lstlisting}[breaklines,mathescape]
n ~ poisson(6).
pos(P):0 ~ uniform(0,100) $\leftarrow$ n~=N, between(1,N,P).
pos(P):t+1 ~ gaussian(X+3, $\Sigma$) $\leftarrow$ pos(P):t~=X.
left(O1,O2):t ~ finite([0.99:true, 0.01:false]) $\leftarrow$
    pos(O1):t~=P1, pos(O2):t~=P2, P1<P2.
\end{lstlisting}
}
\end{addmargin}

The first  rule  states that the number of objects n in the world is distributed according to a Poisson distribution with mean $6$. 
The second rule  states that the position of the n objects, which are identified by a number P between 1 and n, are distributed according to a uniform distribution between 0 and 100. Here, the notation \lstinline[columns=fixed]{n~=N} means that the logical variable N takes the value of our random variable n. The label $0$ (resp. $t$) in the program denotes the point in time. So, \lstinline[columns=fixed]{pos(P):0} denotes the position of object P at time 0. Next, the program describes how the position evolves over time: at each time step the object moves three units of length, giving it a velocity of $3\text{ }[length]/[time]$. Finally, the example program defines the \lstinline[columns=fixed]{left} predicate, through which a relationship between each object is introduced at each time step. DDC then allows for querying this program through its builtin predicate:
\begin{addmargin}{1em}
{\small
\begin{lstlisting}[breaklines,mathescape]
query((left(1,2):t~=true, pos(1):t>0), Probability)
\end{lstlisting}
}
\end{addmargin}
\lstinline[columns=fixed]{Probability} in the second argument unifies with the probability of object \lstinline[columns=fixed]{1} being to the left of object \lstinline[columns=fixed]{2} and having a positive coordinate position.
\end{example}

A DDC program $\mathbb{P}$ is a set of distributional and/or definite clauses (as in Prolog). A DDC program  defines a probability distribution $p(x)$ over possible worlds $x$.
\begin{example} One possible world of the uncountably many possible worlds encoded by the program in Example~\ref{example:ddc_example}. The sampled number \lstinline{n} determines that $2$ objects exists, for which the ensuing distributional clauses then generate a position and  the \lstinline{left/2} relationship:
\begin{addmargin}{1em}
{\small
\begin{lstlisting}[breaklines,mathescape]
n ~= 2.
pos(1):t ~= 30.5. pos(1):t ~= 63.2.
pos(1):t+1 ~= 32.4. pos(1):t+1 ~= 58.8.
left(1,2):t ~= true. left(2,1):t ~= false.
\end{lstlisting}
}
\end{addmargin}
\end{example}
When performing inference within a specific time step, DDC deploys importance sampling combined with backward reasoning (SLD-resolution), likelihood weighting and Rao-Blackwellization~\citep{Nitti:2016:PLP:2949339.2949375}. Inferring probabilities in the next time given the previous time step is achieved through particle filtering \citep{nitti2013particle}. If the DDC program does not contain any predicates labelled with a time index the program represents a \textit{Distributional Clauses} (DC)~\citep{gutmann2011magic} program, where  filtering over time steps is not necessary.


\subsection{Occlusions}\label{sec:theory_occlusion}
Object occlusion is a challenging problem in visual tracking and a plethora of different approaches exist that tackle different kinds of occlusions; a thorough review of the field is given in~\cite{meshgi2015state}. The authors use three different attributes of an occlusion to categorize it: the \textit{extent} (partial or full occlusion), the \textit{duration} (short or long), and the \textit{complexity} (simple or complex)\footnote{An occlusion of an object is deemed complex if during the occlusion the occluded object considerably changes one of its key characteristics, e.g. position, color, size). An occlusion is simple if it is not complex.}. Another classification of occlusions separates occlusions into \textit{dynamic occlusions}, where objects in the foreground occlude each other and \textit{scene occlusions}, where objects in the background model are located closer to the camera and occlude target objects by being moved between the camera and the target objects\footnote{Further categories exist, we refer the reader to \cite{vezzani2011probabilistic,meshgi2015state}.}.

\citeauthor{meshgi2015state} report that the majority of  research on occlusions in visual tracking has been done on partial, temporal and simple occlusions. Furthermore, they report that none of the approaches examined in the comparative studies of~\cite{smeulders2013visual} and~\cite{wu2013online}, handles either partial complex occlusions or full long complex occlusions. To the best of our knowledge, our previous paper on combining bottom-up anchoring and probabilistic reasoning, constitutes the first tracker that is capable of handling occlusions that are full, long and complex~\citep{persson2019semantic}. This was achieved by declaring a \textit{theory of occlusion} (ToO) expressed as dynamic distributional clauses.
\begin{example}\label{example:toc_handcoded}
An excerpt from the set of clauses that constitute the ToO. The example clause describes the conditions under which an object is considered a potential \lstinline{Occluder} of an other object \lstinline{Occluded}.
\begin{addmargin}{1em}
{\small
\begin{lstlisting}[breaklines,mathescape]
occluder(Occluded,Occluder):t+1 ~ finite(1.0:true) $\leftarrow$
    observed(Occluded):t,
    \+observed(Occluded):t+1,     %not observed in next time step
    position(Occluded):t ~= (X,Y,Z),
    position(Occluder):t+1 ~= (XH,YH,ZH),
    D is sqrt((X-XH)^2+(Y-YH)^2), Z<ZH, D<0.3.
\end{lstlisting}
}
\end{addmargin}
Out of all the potential \lstinline{Occluder}'s the actual occluding object is then sampled uniformly:
\begin{addmargin}{1em}
{\small
\begin{lstlisting}[breaklines,mathescape]
occluded_by(Occluded,Occluder):t+1 $\leftarrow$
    sample_occluder(Occluded):t+1 ~= Occluder.
sample_occluder(Occluded):t+1 ~ uniform(ListOfOccluders) $\leftarrow$
    findall(O, occluder(Occluded, O):t+1, ListOfOccluders).
\end{lstlisting}
}
\end{addmargin}

\end{example}

Declaring a theory of occlusion and coupling it to the anchoring system allows the anchoring system to perform \textit{occlusion reasoning} and to track objects not by directly observing them but by reasoning about relationships that occluded objects have entered with visible (anchored) objects. The idea of declaring a theory of occlusion first appeared in~\cite{nitti2013particle}, where, however, the data association problem was assumed to be solved by using AR tags.

As the anchoring system was not able to handle probabilistic states in our previous work, the theory of occlusion had to describe unimodal probability distributions. In this paper we repair this deficiency (cf. Section~\ref{section:probabilistic_anchoring}). Moreover, the theory of occlusion had to be hand-coded (also the case for~\cite{nitti2013particle}). We replace the hand-coded theory of occlusion by a learned one (cf. Section~\ref{section:learning_ddc}). 

Considering our previous work from the anchoring perspective, our approach is most related to the techniques proposed in~\cite{elfring.et.al-2013}, who introduced the idea of \textit{probabilistic multiple hypothesis anchoring} in order to match and maintain probabilistic tracks of anchored objects, and thus, maintain an adaptable \textit{semantic world model}. From the perspective of how occlusions are handled, \citeauthor{elfring.et.al-2013}'s and our work differs, however, substantially. \citeauthor{elfring.et.al-2013} handle occlusions that are due to \textit{scene occlusion}. Moreover, the occlusions are handled by means of a multiple hypothesis tracker, which is suited for short occlusions rather then long occlusions. The limitations with the use of multiple hypothesis tracking for world modeling, and consequently also for handling object occlusions in anchoring scenarios (as in \cite{elfring.et.al-2013}), have likewise been pointed out in a publication by~\cite{wong.et.al-2015}. \citeauthor{wong.et.al-2015} reported instead the use of a clustering-based \textit{data association} approach (opposed to a tracking-based approach), in order to aggregate a consistent semantic world model from multiple viewpoints, and hence, compensate for partial occlusions from a single viewpoint perspective of the scene.


\section{Anchoring of Objects in Multi-Modal States}
\label{section:anchoring}

\begin{figure}[ht!]
	\begin{center}
		\includegraphics[width=0.98\textwidth]{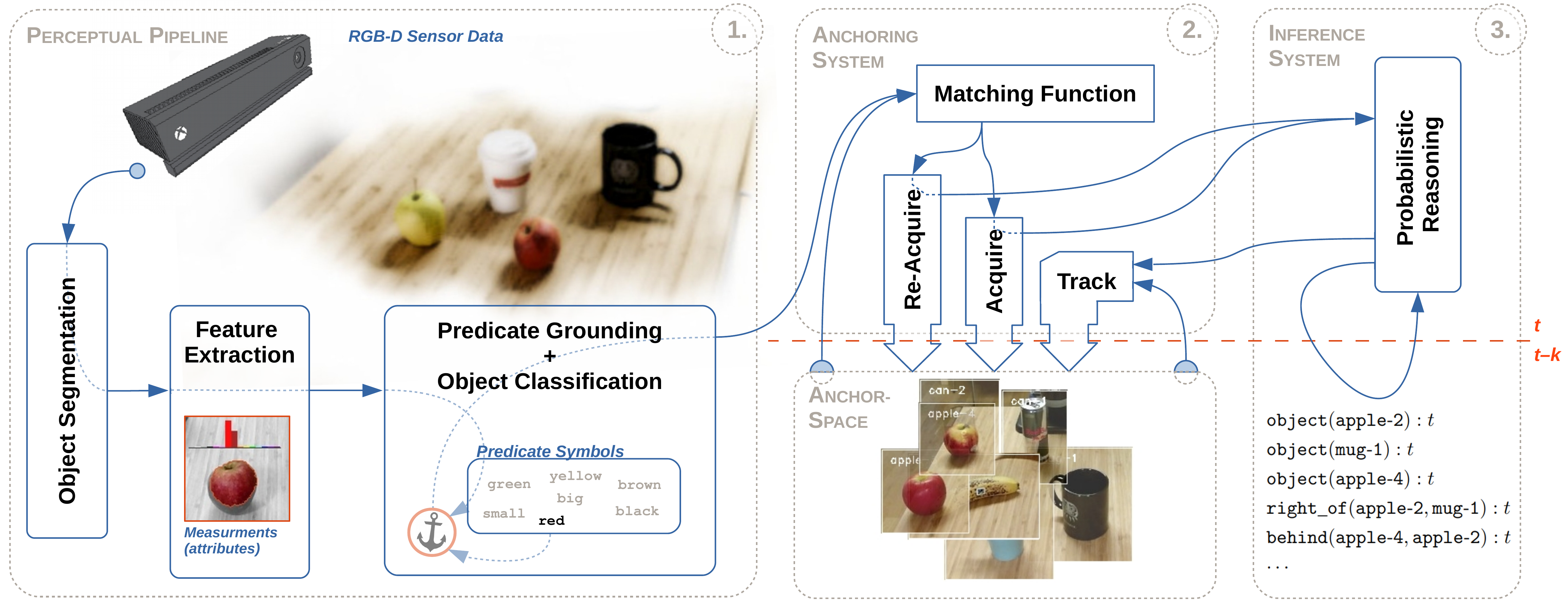}
	\end{center}
	\caption{The overall framework is divided into three basic sub-systems (or modules): \circled{1} an initial \textit{perceptual processing pipeline} for  detecting, segmenting and processing perceived objects, \circled{2} an \textit{anchoring system} for creating and maintaining updated and consistent representations (anchors) of perceived objects, and \circled{3} an \textit{inference system} for aiding the anchoring system and logically tracking  objects in complex dynamic scenes.
	}
	\label{fig:system_overview} 
\end{figure}

In this section, we present a \textit{probabilistic anchoring framework} based on our previous work on conjoining probabilistic reasoning and object anchoring~\citep{persson2019semantic}. An overview of our proposed framework, which is implemented utilizing the libraries and communication protocols available in the Robot Operating System (ROS)\footnote{The code can be found online at: \url{https://bitbucket.org/reground/anchoring}}, can be seen in Figure~\ref{fig:system_overview}. However,  our prior \textit{anchoring system}, seen in Figure~\ref{fig:system_overview}--\circled{2}, was unable to handle probabilistic states of objects. While the \textit{probabilistic reasoning module}, seen in Figure~\ref{fig:system_overview}--\circled{3}, was able to model the position of an object as a probability distribution over possible positions, the anchoring system only kept track of a single deterministic position: the expected position of an object. Therefore, we extend the anchoring notation towards a probabilistic anchoring approach, in order to enable the anchoring system to handle multi-modal probability distributions.

\subsection{Requirements for Anchoring and Semantic Object Tracking}
\label{section:requirements_of_framework}

Before presenting our proposed probabilistic anchoring approach, we first introduce the necessary requirements and assumptions (which partly originate in our previous work,~\cite{persson2019semantic}):

\begin{enumerate}

    \item We assume that unknown anchor representations, $\alpha_y$, are supplied by a \textit{black-box} perceptual processing pipeline, as exemplified in Figure~\ref{fig:system_overview}--\circled{1}. They consist of extracted \textit{attribute measurements} and corresponding grounded \textit{predicate symbols}. We further assume that for each perceptual representation of an object, we have the following attribute measurements: \textit{1)} a \textit{color attribute} ($\phi^{color}_y$), \textit{2)} a \textit{position attribute} ($\phi^{pos}_y$), and \textit{3)} a \textit{size attribute} ($\phi^{size}_y$).
    
    \begin{example}\label{perceptual_example}
    In this paper we use the combined Depth Seeding Network (DSN) and Region Refinement Network (RNN), as presented by \cite{xie2019}, for the purpose of segmenting arbitrary object instances in tabletop scenarios. This two-stage approach leverages both RGB and depth data (given by a Kinect V2 RGB-D sensor), in order to first segment rough initial object masks (based on depth data), followed by a second refinement stage of these object masks (based on RGB data). The resulting output for each segmented object, is then both a \textit{$3{\text -}D$ spatial percept} ($\phi_y^{spatial}$), as well as a \textit{$2{\text -}D$ visual percept} ($\phi_y^{visual}$). For each segmented \textit{spatial percept}, and with the use of the Point Cloud Library (PCL), are both a \textit{position attribute} measured as the $3{\text -}D$ geometrical center, and a \textit{size attribute} measured as the $3{\text -}D$ geometrical bounding box. Similarly, using the Open Computer Vision Library (OpenCV), a \textit{color attribute} is measured as the \textit{discretized color histogram} (in HSV color-space) for each segmented visual percept, as depicted in Figure~\ref{fig:color_attributes}.
    \end{example} 

    \item In order to \textit{semantically categorize objects}, we  assume a Convolutional Neural Network (CNN), such as the GoogLeNet model~\citep{szegedy.et.al-2015}, is available, cf.~\cite{persson.et.al-2017}. In the context of anchoring, the inputs for this model are segmented \textit{visual percepts} ($\pi^{visual}_y$), while resulting object categories, denoted by the predicate $p^{category}_y \in \mathcal{P}$, are given together with the predicted probabilities $\phi^{category}_y$ (cf. Section~\ref{section:background_anchoring}). An example of segmented objects together with the \textit{3-top best object categories}, given by an integrated GoogLeNet model, is illustrated in Figure~\ref{fig:category_classification}. In addition, this integrated model is also used to enhance the traditional \textit{acquire} functionality such that a unique identifier $x$ is generated based on the object category symbol $p^{category}$. For example, if the anchoring system detects an object it has not seen before and classifies it as a \predicate{cup}, a corresponding unique identifier $x = \textnormal{\predicate{cup-4}}$ could be generated (where the $4$ means that this is the forth distinct instance of a \predicate{cup} object perceived by the system).

    \begin{figure}[ht!]
        \centering
        \hfill
        \begin{minipage}[b]{0.46\linewidth}
            \includegraphics[width=\linewidth]{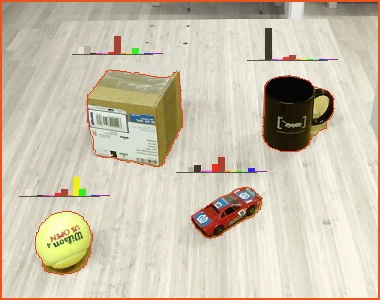}
            \caption{Examples of measured \textit{color attribute} (measured as the \textit{discretized color histogram} over each segmented object).}
            \label{fig:color_attributes}
        \end{minipage} 
        \hfill
        \begin{minipage}[b]{0.46\linewidth}
            \includegraphics[width=\linewidth]{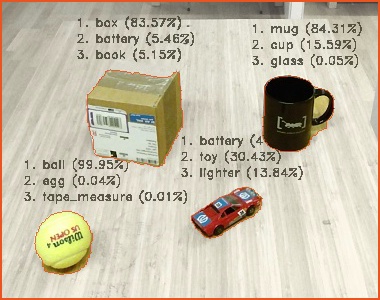}
            \caption{Examples of \textit{semantically categorize objects} (depicted with the \textit{3-top best object categories} for each segmented object).}
            \label{fig:category_classification}
        \end{minipage}
    \end{figure}

    \item We require the presence of a \textit{probabilistic inference system} coupled to the \textit{anchoring system}, as illustrated in Figure~\ref{fig:system_overview}--\circled{3}. The anchoring system is responsible for maintaining objects perceived by the sensory input data and for maintaining the observable part of the world model. Maintained anchored object representations are then treated as observations in the inference system, which uses relational object tracking to infer the state of occluded objects through their relations with perceived objects in the world. This inferred belief of the world is then sent back to the anchoring system, where the state of occluded objects is updated. The feedback-loop between the anchoring system and the probabilistic reasoner results in an additional anchoring functionality~\citep{persson2019semantic}:
    
    \begin{itemize}
    	\item \textit{Track} -- extends the definition of an anchor $\alpha^x$ from time $t-1$ to time $t$. This functionality is directly responding to the state of the probabilistic object tracker, which ensures that the percepts pointed to by the anchor are the adequate perceptual representation of the object, even though the object is currently not perceived.
    \end{itemize}

\end{enumerate}

Even though the mapping between measured attribute values and corresponding predicate symbols is an essential facet of anchoring, we will not cover the \textit{predicate grounding} in further detail in this paper. However, for completeness, we will refer to Figure~\ref{fig:color_attributes} and exemplify that the \textit{predicate grounding relation} of a \textit{color attribute} can, intuitively, be expressed as the encoded correspondence between a specific peek in the color histogram and certain \textit{predicate symbol} (e.g., the symbol \predicate{black} for the \predicate{mug} object). Likewise, a future greater ambition of this work is to establish a practical framework through which the spatial relationships between objects are encoded and expressed using symbolic values, e.g., \textit{object A} is \predicate{underneath} \textit{object B}.


\subsection{Probabilistic Anchoring System}
\label{section:probabilistic_anchoring}

The entry point for the anchoring system, seen in Figure~\ref{fig:system_overview}--\circled{2}, is  a learned \textit{matching function}. This function assumes a bottom-up approach to perceptual anchoring, described in~\cite{loutfi.et.al-2005}, where the system constantly receives candidate anchors and invokes a number of attribute specific matching similarity formulas (i.e., one matching formula for each measured attribute). More specifically, a set of attributes $\Phi_{y}$ of an unknown candidate anchor $\alpha^y_t$  (given at current time $t$) is compared against the set of attributes $\Phi_{x}$ of an existing anchor $\alpha^x_{t-k}$ (defined at time $t-k$) through attribute specific similarity formulas. 
For instance, the similarity between the \textit{positions attributes} $\phi^{pos}_{y}$ of an unknown candidate anchor, and the last updated position $\phi^{pos}_{t-k,x}$ of an existing anchor, is  calculated according to the $L^2$-norm (in $3{\text -}D$ space), which is further mapped to a \textit{normalized similarity value} \citep{blodow.et.al-2010}:

\begin{equation}\label{equation:pos_dist}
  d^{pos}( \phi^{pos}_{t-k,x}, \phi^{pos}_{t,y} ) =  e^{- {L^2}(\phi^{pos}_{t-k,x}, \phi^{pos}_{t,y} ) } 
\end{equation}

Hence, the similarity between two \textit{positions attributes} is given in interval $[0, 1]$, where a value of $1$ is equivalent with perfect correspondence. Likewise, the similarity between two \textit{color attributes} are calculated by the \textit{color correlation}, while the similarity between \textit{size attributes} is calculated according to the \textit{generalized Jaccard similarity} (for further details regarding similarity formulas, we refer to our previous work~\citep{persson2019semantic}). The similarities between the attributes of a known anchor and an unknown candidate anchor are then fed to the learned matching function to determine whether the matching function classifies the unknown anchor to be \textit{acquired} as a new anchor, or \textit{re-acquired} as an existing anchor.

In our prior work on anchoring, the attribute values have always been assumed to be deterministic within a single time step. This assumption keeps the anchoring system de facto deterministic even though it is coupled to a probabilistic reasoning module. We extend the anchoring notation with two distinct specifications of (volatile) attributes: 

\begin{enumerate}
    \item An attribute $\phi_t \in \varphi$ is \emph{deterministic at time $t$} if it takes a single value from the domain $D(\phi_t)$.
    \item An attribute $\phi_t \in \varphi$ is \emph{probabilistic at time $t$} if it is distributed according to a probability distribution $Pr(\phi_t)$ over the domain $D(\phi_t)$ at time step $t$.
\end{enumerate}

Having a probabilistic attribute value $\phi_t$ (e.g., $\phi^{pos}_{t-k,x}$ in Equation~\ref{equation:pos_dist}), means that the similarity calculated with the probabilistic attribute values (e.g., the similarity value $d^{pos}$), will also be probabilistic. Next, in order to use an anchor matching function together with probabilistic similarity values, two  extensions are possible: 1) extend the anchor matching function to accept random variables (i.e., probabilistic similarity values), or 2) retrieve a point estimate of the random variable. 

We chose the second option as this allows us to reuse the anchor matching function learned in~\cite{persson2019semantic} without the additional expense of collecting data and re-training the anchor matching function.
The algorithm to produce the set of matching similarity values that are fed to the anchor matching function is given in Algorithm~\ref{algo:attribute_compare}, where lines~\ref{algo:line:if_prob}-\ref{algo:line:point_estimate} are the extension proposed in this work.

\begin{algorithm}
  \caption{Attribute Compare}
  \label{algo:attribute_compare}
  \begin{algorithmic}[1]
	\Input $\Phi_x$, $\Phi_y \textit{ -- sets of anchor attribute values} $
    \Output $\mathcal{D}_{x,y} \textit{ -- set of matching similarity values} $
	\Function{AttributeCompare}{$\Phi_x$, $\Phi_y$}
      \State $\mathcal{D}_{x,y} \gets \textit{empty set}$
      \ForEach {$\phi_{t,x} \in \Phi_x \textbf{ and } \phi_{t,y} \in \Phi_y$}
      \If {$\phi_{t-1,x}$ is probabilistic} \label{algo:line:if_prob}
      	\State $\mathcal{D}_{x,y} \overset{+}{\leftarrow} \underset{\phi_{t-1,x}}{point\_estimate}  (d( \phi_{t-1,x}, \phi_{t,y})) $\label{algo:line:point_estimate}
        \Else \Comment{deterministic case}
      	\State $\mathcal{D}_{x,y} \overset{+}{\leftarrow} d( \phi_{t-k,x}, \phi_{t,y} )$
      \EndIf
      \EndFor
      \State \Return {$\mathcal{D}_{x,y}$}
    \EndFunction
  \end{algorithmic}
\end{algorithm}

The $point\_estimate$ function in Algorithm~\ref{algo:attribute_compare} (line~\ref{algo:line:point_estimate}) is attribute specific (indicated by the subscript ($\phi_{t-1,x}$)), i.e. we can chose a different point estimation function for \textit{color attributes} than for \textit{position attributes}. An obvious attribute upon which reasoning can be done is the position attribute, for example, in the case of possible occlusions. In other words, we would like to perform probabilistic anchoring while taking into account the probability distribution of an anchor's position. A reasonable goal is then to match an unknown candidate anchor with the most likely anchor, i.e. with the anchor whose position attribute value is located at the highest mode of the probability distribution of the position attribute values. This is achieved by replacing line~\ref{algo:line:point_estimate} in Algorithm~\ref{algo:attribute_compare} with:
\begin{align}
    &\mathcal{F}^{pos}_{x} \leftarrow \Bigg\{  \phi^{pos}_{t-1,x} \Bigg| \frac{\partial Pr(\phi^{pos}_{t-1,x})}{\partial \phi^{pos}_{t-1,x}}=0 \Bigg\} \\
    &\mathcal{D}_{x,y} \overset{+}{\leftarrow} \underset{\phi^{pos} \in \mathcal{F}^{pos} _{x}}{\max}  (d^{pos}(\phi^{pos}, \phi^{pos}_{t,y}))\label{equation:max_pos_dist}
\end{align}
$\mathcal{F}^{pos}_{x}$ is the set of positions situated at the modes of the probability distribution $Pr(\phi_{t-1,x}^{pos})$. In Equation~\ref{equation:max_pos_dist} we take the $\max$ as the co-domain of the position similarity value $d^{pos}$ is in $[0,1]$, where 1 reflects perfect correspondence (cf. Equation~\ref{equation:pos_dist}).

In \cite{persson2019semantic}, we approximated the probabilistic state of the world in the \textit{inference system} (cf. Figure~\ref{fig:system_overview}--\circled{3}) by $N$ particles, which are updated by means of particle filtering. The precise information that is passed from the inference system to the anchoring system is a list of $N$ particles that approximate a (possible) multi-modal belief of the world. More specifically, an anchor $\alpha^x_t$ is updated according to the $N$ particles of possible states of a corresponding object, maintained in the inference system, such that $N$ possible $3{\text -}D$ positions are added to the volatile \textit{position attributes} $\varphi^{pos}_x$. In practice we assume that samples are only drawn around the modes of the probability distribution, which means that we can replace line~\ref{algo:line:point_estimate} of Algorithm~\ref{algo:attribute_compare} with:
\begin{align}
    \mathcal{D}_{x,y} \overset{+}{\leftarrow} \underset{i}{\max}  \left( d^{pos}( \phi^{pos}_{t-1,x,i}, \phi^{pos}_{t,y} ) \right)
    =  \underset{i}{\max} \left( e^{- {L^2}(\phi^{pos}_{t-1,x,i}, \phi^{pos}_{t,y} )  }  \right)
\end{align}
Where $\phi_{t-1,x,i}$ is a sampled position and $i$ ranges from $1$ to the number of samples $N$. 

Performing probabilistic inference in the coordinate space is a choice made in the design of the probabilistic anchoring system. Instead, the probabilistic tracking could also be done in the HSV color space, for instance. In this case, the similarity measure used in Algorithm~\ref{algo:attribute_compare} would have to be adapted accordingly. It is also conceivable to combine the tracking in coordinate space and color space. This introduces, however, the complication of finding a similarity measure that works on the coordinate space and the color space at the same time. A solution to this would be to, yet again, learn this similarity function from data~\citep{persson2019semantic}. 


\section{Learning Dynamic Distributional Clauses}\label{section:learning_ddc}
While several approaches exist in the SRL literature that learn probabilistic relational models, most of them focus on parameter estimation~\citep{sato1995statistical,friedman1999learning,taskar2002discriminative,neville2007relational} and structure learning
 has  been restricted to discrete data.  Notable 
 exceptions include 
the recently proposed hybrid relational formalism by~\cite{ravkic2015learning}, which learns relational models in a discrete-continuous domain but has not been applied to dynamics or robotics,
and the 
related approach of~\cite{nitti2016learning}, where a relational tree learner DDC-TL  learns both the structure and the parameters of distributional clauses. DDC-TL has been evaluated on learning action models (pre- and post-conditions) in a robotics setting from before- and after-states of executing the actions. However, there were several limitations of the approach.
It simplified perception by resorting to AR tags for identifying the objects, it did not consider occlusion, and it could not deal with uncertainty or noise in the observations.

A more general approach to learning distributional clauses, extended with {\em statistical models}, is being proposed in~\cite{kumar2020learning}\footnote{\url{https://github.com/niteshroyal/DreaML}}. Such a statistical model relates continuous variables in the body of a distributional clause to parameters of the distribution in the head of the clause. The approach simultaneously learns the structure and parameters of (non-dynamic) distributional clauses,  and estimates the parameters of the statistical  model  in clauses. A DC program consisting of multiple distributional clauses  is capable of expressing intricate probability distributions over discrete and continuous random variables. A further shortcoming of DDC-TL (also tackled by \citeauthor{kumar2020learning}) is the inability of learning in the presence of background knowledge --- that is, additional (symbolic) probabilistic information about objects in the world and relations (such as spatial relations) among the objects that the learning algorithm should take into consideration.

However, until now, the approach presented in~\cite{kumar2020learning} has only been applied to the problem of autocompletion of relational databases by learning a (non-dynamic) DC program. We now demonstrate with an example of how this general approach can also be applied for learning dynamic distributional clauses in a robotics setting.
A key novelty in the context of perceptual anchoring is that we learn
a DDC program that allows us to reason about occlusions.

\begin{example}\label{example: occluded_by}
Consider again a scenario where objects might get fully occluded by other objects.
We would now like to learn the ToO that describes whether an object is occluded or not given multiple observations of the before and after state.
In DDC we represent
observations through
facts as follows
\begin{addmargin}{1em}
{\small
\begin{lstlisting}[breaklines,mathescape]
pos(o1_exp1):t ~= 2.3.
pos(o1_exp1)t+1 ~= 9.3.
pos(o2):t ~= 2.2.
pos(o2):t+1 ~= 9.2.
occluded_by(o1_exp1,o2_exp1):t+1.
pos(o3_exp1):t ~= 8.3.
$\text{\vdots}$
\end{lstlisting}
}
\end{addmargin}
For the sake of clarity, we have considered only one-dimensional positions in this example.

Given the data in form dynamic distributional clauses, we are now interested in learning the ToO instead of relying on a hand-coded one, as in Example~\ref{example:toc_handcoded}. An excerpt from the set of clauses that constitute a learned ToO is given below. As in Example~\ref{example:toc_handcoded}, the clause describe the circumstances under which an object (\lstinline{Occluded}) is potentially occluded by an other object (\lstinline{Occluder}).

\begin{addmargin}{1em}
{\small
\begin{lstlisting}[breaklines,mathescape]

occluder(Occluded,Occluder):t+1 ~ finite(1.0:false) $\leftarrow$ 
    occluded_by(Occluded,Occluder):t,
    observed(Occluded):t+1.
occluder(Occluded,Occluder):t+1 ~ finite(0.92:true,0.08:false) $\leftarrow$
    occluded_by(Occluded,Occluder):t,
    \+observed(Occluded):t+1.
occluder(Occluded,Occluder):t+1 ~ finite(P1:true,P2:false) $\leftarrow$                
    \+occluded_by(Occluded,Occluder):t,
    \+observed(Occluded):t+1,
    distance(Occluded,Occluder):t~=Distance, 
    logistic([Distance],[-16.9,0.8],P1), 
    P2 is 1-P1.
\end{lstlisting}
}
\end{addmargin}
Note that, in the second but last line of the last clause above the arbitrary threshold on the \lstinline{Distance} is superseded by a learned statistical model, in this case a logistic regression, which maps the input parameter \lstinline{Distance} to the probability \lstinline{P1}:
\begin{align}
    \text{\lstinline{P1}}=\frac{1}{1+ e^{16.9{\times}\text{\lstinline{D}-0.8}}}
\end{align}
Replacing the hand-coded \lstinline{occluder} rule with the learned one in the theory of occlusion allows us to track occluded objects with a partially learned model of the world.

\end{example}

In order to learn dynamic distributional clauses, we first map the predicates with subscripts that refer to the current time step \lstinline{t} and the next time step \lstinline{t+1} to standard predicates, which gives us an input DC program. For instance, we map \lstinline{pos(o1_exp1):t} to \lstinline{pos_t(o1_exp1)}, and \lstinline{occluder(o1_exp1,o2_exp2):t+1} to \lstinline{occluder_t1(o1_exp1,o2_exp2)}. The method introduced in~\citep{kumar2020learning} can now be applied for learning distributional clauses for the target predicate \lstinline{occluder_t1(o1_exp1,o2_exp2)} from the input DC program. 

\begin{figure}[ht!]
	\begin{center}
		\includegraphics[height=8.0cm]{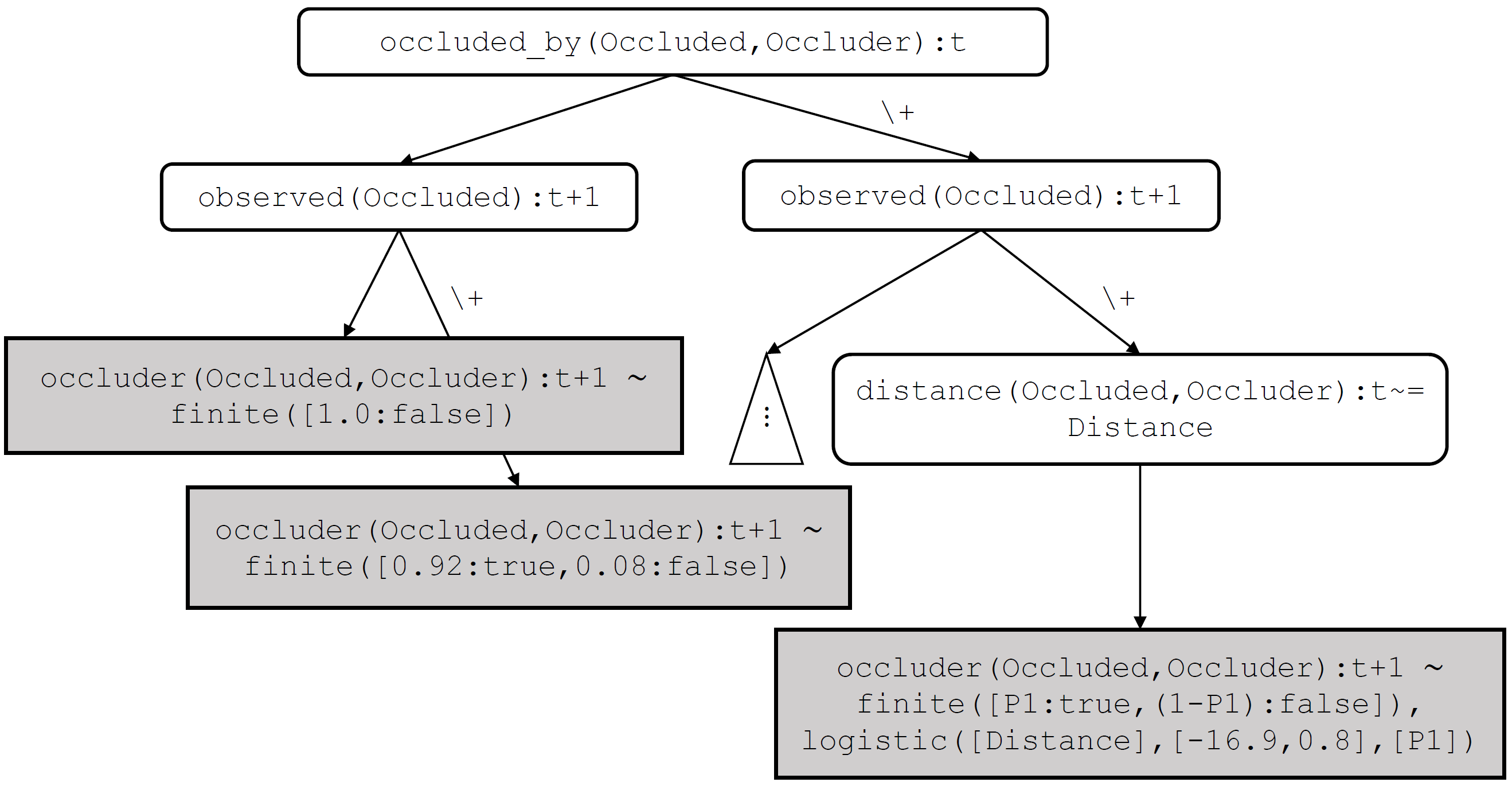}
	\end{center}
	\caption{A distributional logic tree that represents learned clauses for the target \lstinline{occluder(Occluded,Occluder):t+1}. The leftmost path corresponds to the first clause, the rightmost path corresponds to the last clause for \lstinline{occluder(Occluded,Occluder):t+1} in Example~\ref{example: occluded_by}. Internal nodes such as \lstinline{occluder(Occluded,Occluder):t} and \lstinline{observed(Occluded):t+1} are discrete features, whereas, internal nodes such as \lstinline{distance(Occluded,Occluder):t+1\~=Distance} is a continuous feature. 
	}\label{fig:trees} 
\end{figure}

Clauses for the target predicate are learned by inducing a distributional logic tree. An example of such a tree is shown in Figure \ref{fig:trees}. The key idea is that the set of clauses for the same target predicate are represented by a distributional logic tree, which satisfies the mutual exclusiveness property of distributional clauses. This property states that if there are two distributional clauses defining the same random variable, their bodies must be mutually exclusive. Internal nodes of the tree correspond to atoms in the body of learned clauses. A leaf node corresponds to a distribution in the head and to a statistical model in the body of a learned clause. A path beginning at the root node and proceeding to a leaf node in the tree corresponds to a clause. Parameters of the distribution and the statistical model of the clause are estimated by maximizing the expectation of the log-likelihood of the target in partial possible worlds. The worlds are obtained by proving all possible groundings of the clause in the input DC program. The structure of the induced tree defines the structure of the learned clauses. The approach requires declarative bias to restrict the search space while inducing the tree. 

In summary, the {\em input} to the learning algorithm is a DC program consists of
\begin{itemize}
        \item background knowledge, in the form of DC clauses;
        \item observations, in the form of DC clauses --- these constitute the training data;
        \item the declarative bias, which is necessary to specify the hypothesis space of the DC program~\citep{ade1995declarative};
        \item the target predicates for which clauses should be learned.
\end{itemize}
The {\em output} is:
\begin{itemize}
    \item a set of DC clauses represented as a tree for each target predicate specified in the input.
\end{itemize}

Once the clauses are learned, predicates are mapped back to predicates with subscripts to obtain dynamic distributional clauses. For instance, \lstinline{occluder_t1(Occluded,Occluder)} in the learned clauses is mapped back to \lstinline{occluder(Occluded,Occluder):t+1}.

\begin{figure}
    \begin{center}
		\includegraphics[width=0.98\textwidth]{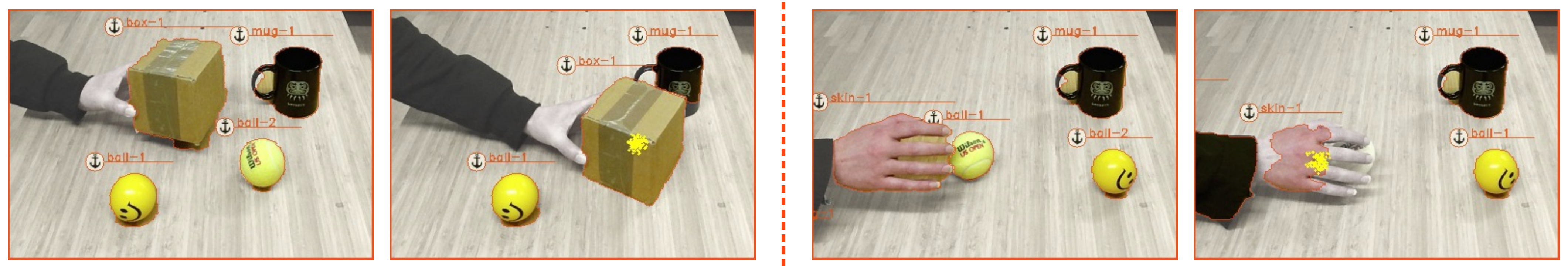}
	\end{center}
    \caption{
        Depicted are two training points in the data set that were used to learn the transition rule of an object to another object. The panels on the left show a \predicate{ball} that is being occluded by a \predicate{box}, and on the right, the same \predicate{ball} that is being grabbed by a hand (or a \predicate{skin} object, as we have only trained our used GoogLeNet model to recognize general human skin objects instead of particular human body parts, cf. Section~\ref{section:requirements_of_framework}). The plotted dots on top of the occluding object represent samples drawn from the probability distribution of the occluded object, in other words the object that is labeled in the data set to transition into the occluding counterpart.
    }
    \label{fig:data_collection}
\end{figure}

The data used for the learning of the theory of occlusion consists of training points of before-after states of two kinds. The first kind are pairs describing a transition of an objection from being observed to being occluded. The second kind of data pairs describe an object being occluded in the current state as well as in the next state. Examples of two raw data points for the first kind can be seen in Figure~\ref{fig:data_collection}. The processed data that was fed to the distributional clauses learner is available online\footnote{\url{https://bitbucket.org/reground/anchoring/downloads/}} as well as models with the learned theory of occlusion\footnote{\url{https://bitbucket.org/reground/anchoring}}.

\section{Evaluation}\label{section:experiments}

A probabilistic anchoring system that is coupled to an inference system (cf. Section~\ref{section:probabilistic_anchoring}) is comprised of several interacting components. This turns the evaluation of such a combined framework, with many integrated systems, into a challenging task. We, therefore, evaluate the integrated framework as a whole on representative scenarios (videos of which are online\footnote{\url{https://vimeo.com/manage/folders/1365568}}) that demonstrate our proposed extensions to perceptual anchoring. In Section~\ref{section:eval_multimodal}, we demonstrate how the extended anchoring system can handle probabilistic multi-modal states (described in Section~\ref{section:anchoring}). In Sections~\ref{section:unimodal} and~\ref{section:transitive}, we show that semantic relational object tracking can be performed with the probabilistic logic rules (in form of a DDC program) instead of handcrafted ones.


\subsection{Multi-Modal Occlusions}\label{section:eval_multimodal}

\begin{figure}[ht!]
	\begin{center}
		\includegraphics[width=0.98\textwidth]{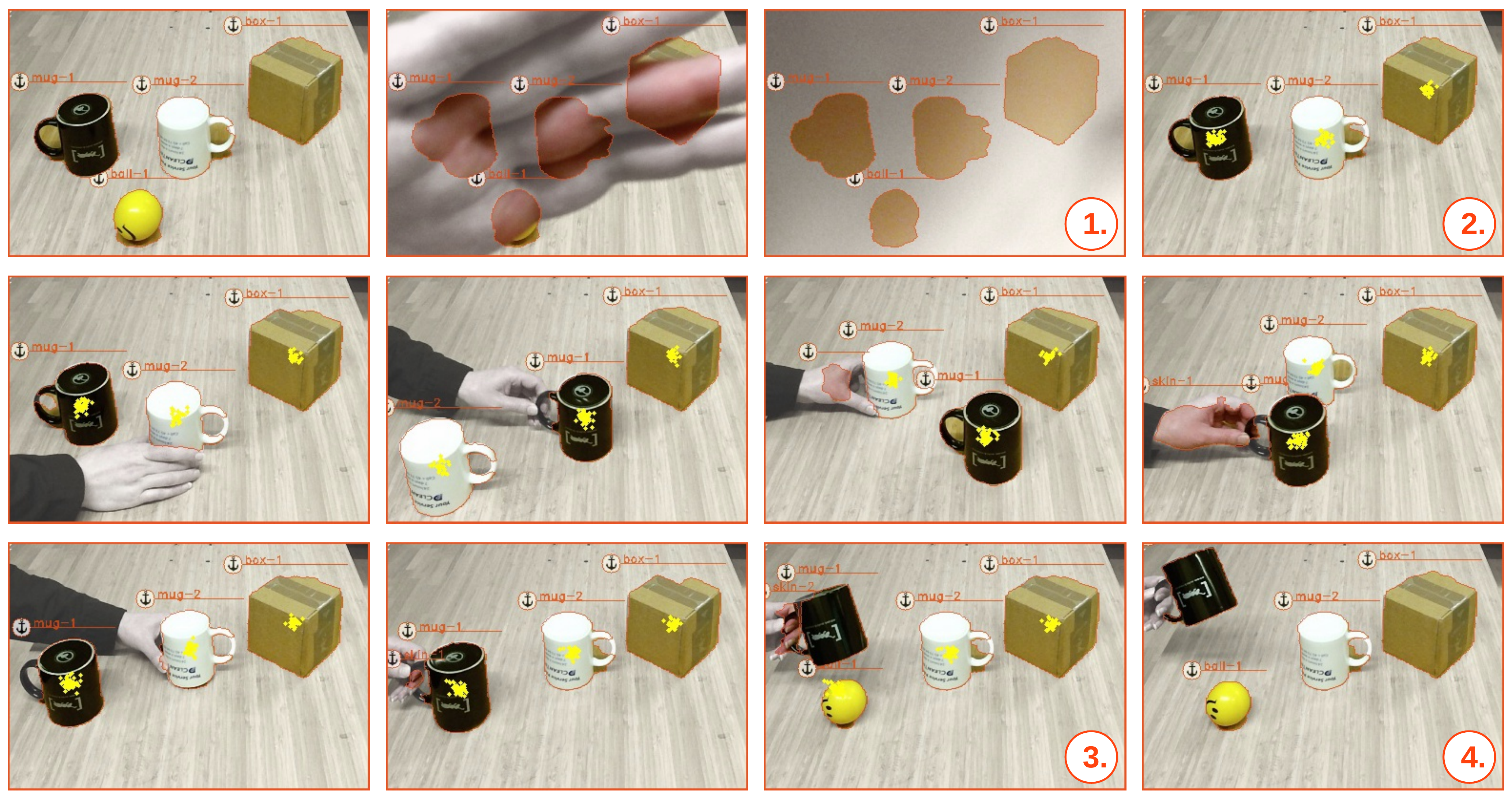}
	\end{center}
	\caption{
	    Screen-shots captured during the execution of a scenario where the stream of sensor data is obscured. Visually perceived anchored objects are symbolized by a unique anchor identifiers (e.g., \predicate{mug-1}), while occluded hidden objects are depicted by plotted particles that represent possible positions of the occluded object in the inference system. The screenshots illustrate a scenario where the RGB-D sensor is covered and a  \predicate{ball} is hidden under either one of three larger objects. These larger objects are subsequently shuffled around before the whereabouts of the hidden \predicate{ball} is revealed.
    }
    \label{fig:multi_modal}
\end{figure}

We present the evaluation in the form of screenshots captured during the execution of a scenario where we obscure the stream of sensor data. We start out with three larger objects (two \predicate{mug} objects and one \predicate{box} object), and one smaller \predicate{ball} object. During the occlusion phase, seen in Figure~\ref{fig:multi_modal}--\circled{1}, the RGB-D sensor is covered by a human hand and the smaller \predicate{ball} is hidden underneath one of the larger objects. In Figure~\ref{fig:multi_modal}--\circled{1}, it should also be noted that the anchoring system preserves the latest update of the objects, which is here illustrated by the outlined contour of each object.
At the time that the sensory input stream is uncovered, and there is no longer any visual perceptual input of the \predicate{ball} object, the system can only speculate about the whereabouts of the missing object. Hence, the belief of the \predicate{ball}'s position becomes a multi-modal probability distribution, from which we draw samples, as seen in Figure~\ref{fig:multi_modal}--\circled{2}. At this point, we are, however, able to track the smaller \predicate{ball} through its probabilistic relationships with the other larger objects. During all the movements of the larger objects, the probabilistic inference system manages to track the modes of the probability distribution of the position of the smaller \predicate{ball}. The probability distribution for the position of the smaller \predicate{ball} (approximated by $N$ samples) is continuously fed back to the anchoring system. Consequently, once the hidden \predicate{ball} is revealed and reappears in the scene, as seen in Figures~\ref{fig:multi_modal}--\circled{3} and~\ref{fig:multi_modal}--\circled{4}, the \predicate{ball} is correctly \textit{re-acquired} as the initial \predicate{ball-1} object. This would not have been possible with a non-probabilistic anchoring approach.

\subsection{Uni-Modal Occlusions with Learned Rules}\label{section:unimodal}
\label{section:uni-modal_occlusion}

\begin{figure}[ht!]
    \begin{center}
		\includegraphics[width=0.98\textwidth]{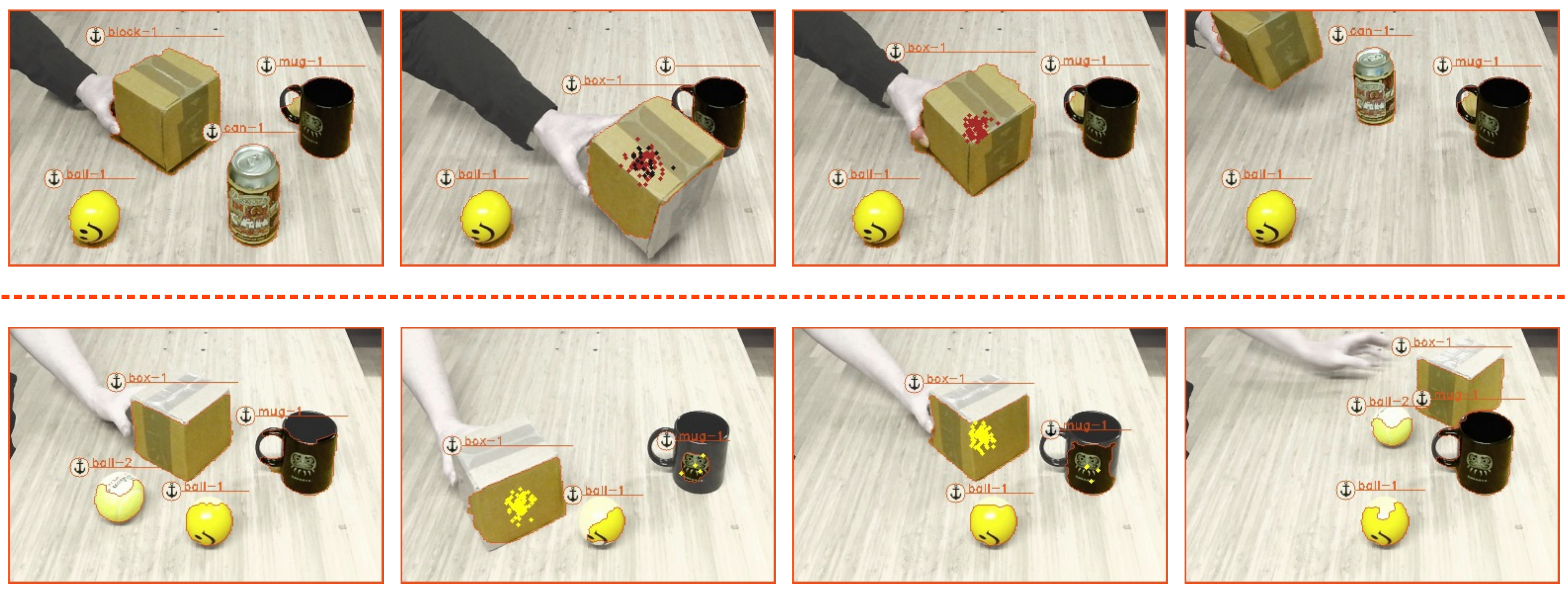}
	\end{center}
    \caption{
        The two scenario show how a learned ToO is used to perform semantic relational object tracking. In both scenarios, an object is occluded by a \predicate{box} and successfully tracked before the occluded object is being revealed and again \textit{re-acquired} as the same initial object.
    }
    \label{fig:uni_modal}
\end{figure}

The conceptually easiest ToO is one that describes the occlusion of on object by an other object. Using the method described in Section~\ref{section:learning_ddc}, we learned such a ToO, which we demonstrate in Figure~\ref{fig:uni_modal}. Shown are two scenarios. In the one in the upper row the a \predicate{can} gets occluded by a \predicate{box} --- shown in the second screenshot. The \predicate{can} is subsequently tracked through its relation with the observed \predicate{box} and successfully re-anchored as \predicate{can-1} once it is revealed. Note that in the second screenshot, the \predicate{mug} is also briefly believed to be hidden under \predicate{box}, shown through the \predicate{black} dots, as is the \predicate{mug} is temporally obscured behind the \predicate{box} and not observed by the vision system. However, once the \predicate{mug} is again observed the black dots disappear.

In the second scenario, we occlude one of two \predicate{ball} objects with a \predicate{box} and track the \predicate{ball} again through its relation with the \predicate{box}. Note that some of the probability mass accounts for the possibility for the occluded \predicate{ball} to be occluded by the \predicate{mug}. This is due to the fact that the learned rule is probabilistic. 

In both scenarios, we included background knowledge that specifies that a \predicate{ball} cannot be the an occluder of an object (it does not \textit{afford} to be the occluder). This is also why we see a probability mass of the occluded \predicate{ball} at the \predicate{mug}'s location and not at the observed \predicate{ball}'s location in the second scenario.


\subsection{Transitive Occlusions with Learned Rules}
\label{section:transitive}

Learning (probabilistic) rules, instead of a black-box function, has the advantage that a set of rules can easily be extended with further knowledge. For example, if we would like the ToO to be recursive, i.e., objects can be occluded by objects that are themselves occluded, we simply have to add the following rule to the DDC program describing the theory of occlusion:  
\begin{addmargin}{1em}
{\small
\begin{lstlisting}[breaklines,mathescape]
occluded_by(Occluded,Occluder):t+1 $\leftarrow$
    occluded_by(Occluded,Occluder):t,
    \+observed(Occluded):t+1,
    \+observed(Occluder):t+1,
    occluded_by(Occluder,_):t+1.
\end{lstlisting}
}
\end{addmargin}

\begin{figure}[ht!]
    \begin{center}
		\includegraphics[width=0.98\textwidth]{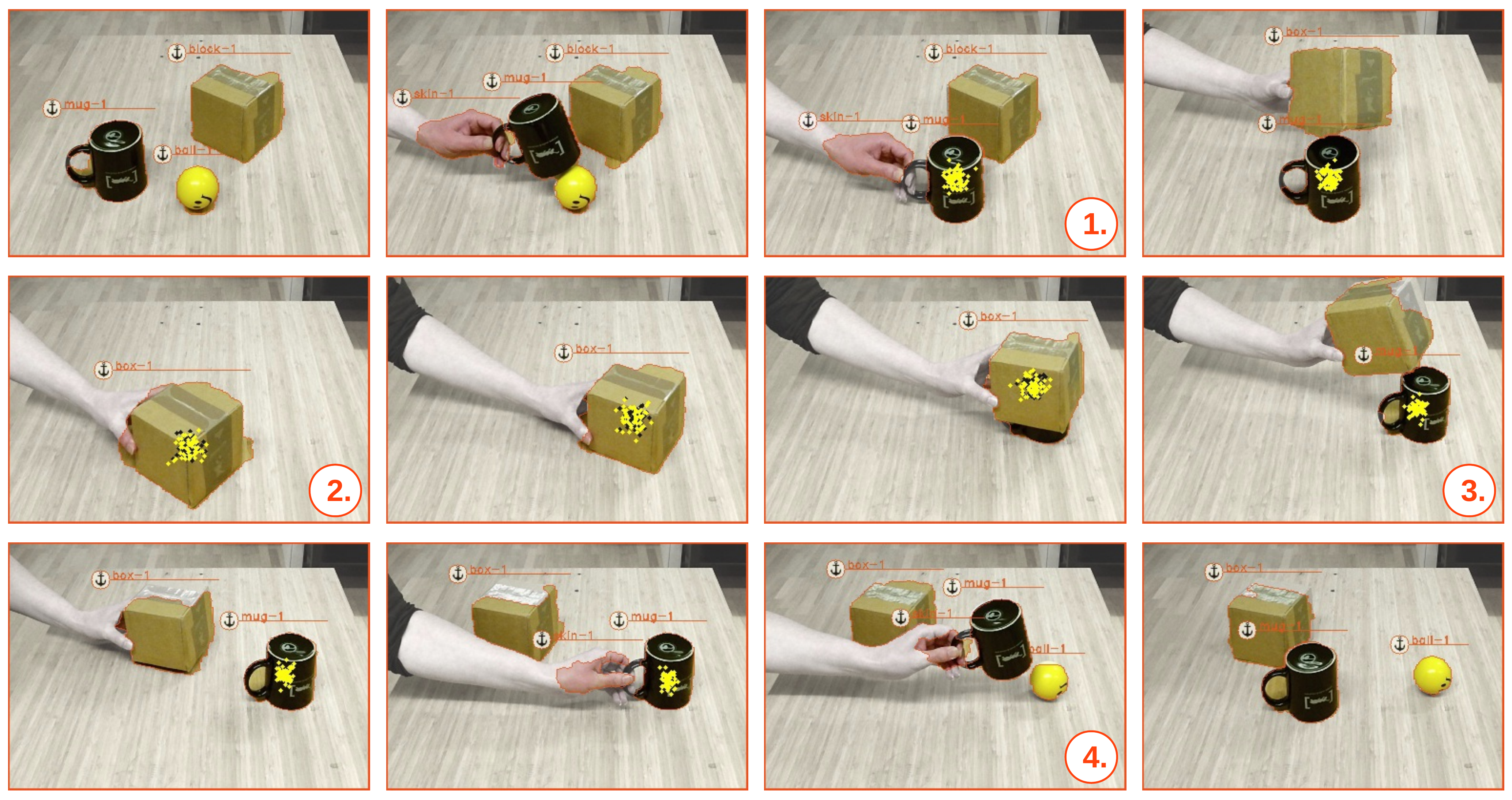}
	\end{center}
    \caption{
        A scenario that demonstrates transitive occlusions based on learned rules for handling the theory of occlusions. First the \predicate{ball} is occluded by the \predicate{mug} (indicated by the yellow dots) and subsequently the \predicate{mug} is occluded in turn by the \predicate{box} (indicated by the black dots). Once the \predicate{mug} is observed again the \predicate{ball} is still believed to be occluded by the \predicate{mug}.
    }
    \label{fig:transitive}
\end{figure}

Extending the ToO from Section~\ref{section:unimodal} with the above rule, enables the anchoring system to handle recursive occlusions. We demonstrate such a scenario in Figure~\ref{fig:transitive}. Initially, we start this scenario with a \predicate{ball}, a \predicate{mug} and a \predicate{box} object (which in the beginning is miss-classified as \predicate{block} object, cf. Figure~\ref{fig:category_classification}). In the first case of occlusion, seen in Figures~\ref{fig:transitive}--\circled{1}, we have the same type of uni-modal occlusion as described in the previous Section~\ref{section:uni-modal_occlusion}, where the \predicate{mug} occludes the \predicate{ball} and, subsequently, triggers the learned relational transition (where plotted \predicate{yellow} dots represent samples drawn from the probability distribution of the occluded \predicate{ball} object). In the second recursive case of occlusion, seen in Figure~\ref{fig:transitive}--\circled{2}, we proceed by also occluding the \predicate{mug} with the \predicate{box}. Above rule administers this \textit{transitive occlusion} --- triggered when the \predicate{ball} is still hidden underneath the \predicate{mug} and the \predicate{mug} is occluded by the \predicate{box}. This is illustrated here by both \predicate{yellow} and \predicate{black} plotted dots that represent samples drawn from the probability distributions of occluded \predicate{mug} and the transitively occluded \predicate{ball} object, respectively. Consequently, once the \predicate{box} is moved, both the \predicate{mug} and the \predicate{ball} are tracked through the transitive relation with the occluding \predicate{box}. Reversely, it can be seen, in Figure~\ref{fig:transitive}--\circled{3}, that once the \predicate{mug} object is revealed the object is correctly \textit{re-acquired} as the same \predicate{mug-1} object, while the relation between the \predicate{mug} and the occluded \predicate{ball} object is still preserved. Finally, as the \predicate{ball} object is revealed, in Figure~\ref{fig:transitive}--\circled{4}, it can be also seen that the object is, likewise, correctly \textit{re-acquired} as the same \predicate{ball-1} object.

\section{Conclusions \& Future Work}
\label{section:conclusions}
We have presented a two-fold extension to our previous work on semantic world modelling~\citep{persson2019semantic}, where we proposed an approach to couple an anchoring system to an inference system. Firstly, we extended the notions of perceptual anchoring towards the probabilistic setting by means of probabilistic logic programming. This allowed us to maintain a multi-modal probability distribution of the positions of objects in the anchoring system and to use it for matching and maintaining objects at the perceptual level --- thus, we introduce probabilistic anchoring of objects either directly perceived by the sensory input data or logically inferred through probabilistic reasoning. We illustrated the benefit of this approach with the scenario in Section~\ref{section:eval_multimodal}, which the anchoring system was able to resolve correctly only due to its ability of maintaining a multi-modal probability distribution. This also extends an earlier approach to relational object tracking~\citep{nitti2014relational}, where the symbol-grounding problem was solved by the use of AR tags.

Secondly, we have deployed methods from statistical relational learning to the field of anchoring. This approach allowed us to learn, instead of handcraft, rules needed in the reasoning system. A distinguishing feature of the applied rule learner~\citep{kumar2020learning} is its ability to handle both continuous and discrete data. We then demonstrated that combining perceptual anchoring and SRL is also feasible in practice by performing relational anchoring with a learned rule (demonstrated in Section~\ref{section:uni-modal_occlusion}). This scenario did also exhibit a further strength of using SRL in anchoring domains, namely that the resulting system becomes a highly modularizable system. In our evaluation, for instance, we were able to integrate an extra rule into the ToO, which enabled us to resolve recursive occlusions (described in Section~\ref{section:transitive}).

A possible future direction would be to exploit how anchored objects and their spatial relationships, tracked over time, facilitate the learning of both the function of objects, as well as object affordances~\citep{kjellstrom.et.al-2011,moldovan2012learning,koppula.et.al-2013,koppula&saxena-2014}.
Through the introduction of a probabilistic anchoring approach, together with the learning of the rules that express the relation between objects, we have presented a potential framework for future studies of spatial relationship between objects, e.g., the spatial-temporal relationships between objects and human hand actions to learn the function of objects (cf.~\cite{kjellstrom.et.al-2011,moldovan2012learning}). Such a future direction would tackle a similar question, currently discussed in the neural-symbolic community~\citep{garcez2019neural}, namely how to propagate back symbolic information to  sub-symbolic representations of the world. A recent piece of work that combines SRL and neural methods is, for instance, \cite{manhaeve2018deepproblog}.

Another aspect of our work that deserves future investigation is probabilistic anchoring, in itself. With the approach presented in this paper we are merely able to perform MAP inference. In order to perform full probabilistic anchoring, one would need to render the anchor matching function itself fully probabilistic, i.e. the anchor matching function would need to take as arguments random variables and again output probability distributions instead of point estimates --- ideas borrowed from multi-hypothesis anchoring~\citep{elfring.et.al-2013} might, therefore, be worthwhile to consider for future work.





\section*{Author Contributions}
PZ and AP outlined the extension of the framework to include probabilistic properties and multi-modal states
PZ and NK  integrated SRL with perceptual anchoring. PZ, AP and NK performed the experimental evaluation.
AL and LD have developed the notions and the ideas in the paper together with the other authors. PZ, NK, AP, AL and LD have all contributed to the text.


\section*{Conflict-of-Interest Statement}
The authors declare that the research was conducted in the absence of any commercial or financial relationships that could be construed as a potential conflict of interest.









\bibliographystyle{frontiersinSCNS_ENG_HUMS} 
\bibliography{references}

\end{document}